\documentclass[11pt]{article}

\usepackage{eamt18}
\usepackage{times}
\usepackage{url}
\usepackage{latexsym}
\usepackage[small,bf]{caption}
\usepackage{color}
\usepackage{xcolor} 
\usepackage[estonian, english]{babel}
\usepackage[utf8]{inputenc}

\title{Multi-Domain Neural Machine Translation}

\author{Sander Tars \and Mark Fishel\\
  Institute of Computer Science\\
  University of Tartu, Estonia\\
  {\tt tarssander1@gmail.com, fishel@ut.ee}}

\date{}

\begin{document}
\maketitle
\begin{abstract}
We present an approach to neural machine translation (NMT) that supports multiple domains in a single model and allows switching between the domains when translating.
The core idea is to treat text domains as distinct languages and use multilingual NMT methods to create multi-domain translation systems; we show that this approach results in significant translation quality gains over fine-tuning. We also explore whether the knowledge of pre-specified text domains is necessary; turns out that it is after all, but also that when it is not known quite high translation quality can be reached, and even higher than with known domains in some cases.
\end{abstract}

\section{Introduction}
Data-driven machine translation (MT) systems depends on the text domain of their training data. In a typical in-domain MT scenario the amount of parallel texts from a single domain is not enough to train a good translation system, even more so for neural machine translation \nocite{bahdanau} (NMT; Bahdanau et al., 2014); thus models are commonly trained on a mixture of parallel texts from different domains and then later fine-tuned to in-domain texts \cite{finetuning}.

In-domain fine-tuning has two main shortcomings: it depends on the availability of sufficient amounts of in-domain data in order to avoid overfitting and it results in degraded performance for all other domains. The latter means that for translating multiple domains one has to run an individual NMT system for each domain.

In this work we treat text domains as distinct languages: for example, instead of English-to-Estonian translation we see it as translating English news to Estonian news. We test two multilingual NMT approaches \cite{Johnson:17,Tiedemann:17} in a bilingual multi-domain setting and show that both outperform single-domain fine-tuning on all the text domains in our experiments.

However, this only works when the text domain is known both when training and translating. In some cases the text domain of the input segment is unknown -- for example, web MT systems have to cope with a variety of text domains. Also, some parallel texts do not have a single domain while they are either a mix of texts from different sources (like crawled corpora) or naturally constitute a highly heterogeneous mix of texts (like subtitles or Wikipedia articles).

We address these issues by replacing known domains with automatically derived ones. At training time we cluster parallel sentences and then applying the multi-domain approach to these clusters. When translating, the input segments are classified as belonging to one of these clusters and translated with this automatically derived information.

In the following we review related work in Section~\ref{sctRelWork}, then present our methodology of multi-domain NMT and sentence clustering in Section~\ref{sctModels}. After that, we describe our experiments in Sections~\ref{sctExp1} and \ref{sctExp2} and discuss the results in Section~\ref{sctAnalysis}. Section~\ref{sctConclusions} concludes the paper.

\section{Related Work}
\label{sctRelWork}

The baseline to which we compare our work is fine-tuning NMT systems to a single text domain \cite{finetuning}. There, the NMT system is first trained on a mix of parallel texts from different domains and then fine-tuned via continued training on just the in-domain texts. The method shows improved performance on in-domain test data but degrades performance on other domains.



In \cite{politeness_sennrich} the NMT system is parametrized with one additional input feature (politeness), which is included as part of the input sequence, similarly to one of our two approaches (in our work -- the domain tag approach). However, their goal is different from ours.

In \cite{domain_control_kobus} additional word features are used for specifying the text domain together with the same approach as \cite{politeness_sennrich}. Although both methods overlap with the first part or our work (domain features and domain tags), they only test these methods on pre-specified domains, while we include automatic domain clustering and identification. Also, they use in-domain trained NMT systems as baselines even for small parallel corpora and do experiments with a different NMT architecture. Finally, their results show very modest improvements, while in our case the improvements are much greater.

Other approaches also define a mixture of domains, for example \cite{effective_mixing_britz,topic_aware_chen}. However, both define custom NMT methods and also limit the experiments to the cases where the text domain is known.

\section{Methodology}
\label{sctModels}

In the following we describe two different approaches to treating text domains as distinct languages and using multi-lingual methods, resulting in multi-domain NMT models. The first approach is inspired by Google's multilingual NMT ~\cite{Johnson:17} and the second one by the cross-lingual language models~\cite{Tiedemann:17}.
Then we describe our methods of unsupervised domain segmentation used in our experiments in comparison with the pre-specified text domains.

\subsection{Domain as a Tag}

The first approach is based on \cite{Johnson:17}. Their method of multilingual translation is based on 
training the NMT model on data from multiple language pairs, while appending a token specifying the target language to the beginning of the source sequence. No changes to the NMT architecture are required with this approach. They show that the method improves NMT for all languages involved; as an additional benefit, there is no increase in the number of parameters, since all language pairs are included in the same model.

We adapt the language tag approach to text domains, appending the domain ID to each source sentence; thus, for instance, ``How you doin' ?'' from OpenSubtitles2016
\cite{opsubs}
becomes ``\_\_OpenSubs How you doin' ?''.

The described method has two advantages. Firstly, it is independent of the NMT architecture, and scaling to more domains means simply adding data for these domains. We can assign a domain to each sentence pair of the training set sentence pair, or set the domain to ``other'' for sentences whose domain we cannot or do not want to identify.

Secondly, in a multilingual NMT model, all parameters are implicitly shared by all the language pairs being modeled. This forces the model to generalize across language boundaries during training. It is observed that when language pairs with little available data and language pairs with abundant data are mixed into a single model, translation quality on the low resource language pair is significantly improved.

We expect this to be even more useful for text domains. Traditional tuning to a low-resource domain, or for any specific domain for that matter, would result in a likely over-fitting to that domain. Our approach, where all parameters are shared, learns target domain representations without harming other domains' results while maintaining the ability to generalize also on in-domain translation, because little to no over-fitting will be caused. Furthermore, since domains are much more similar than languages, we expect the parameter sharing to have a stronger effect.

\subsection{Domain as a Feature}

The second approach is based on \cite{Tiedemann:17} for continuous multilingual language models. The authors propose to use a single RNN model with language vectors that indicate what language is used. As a result each language gets its own embedding, thus ending up with a language model with a predictive distribution $p(x_t\vert x_{1...t-1}, l)$ which is a continuous function of the language vector $l$.

In our approach the same idea is implemented via word features of Nematus \cite{Sennrich:17}, 
with their learned embeddings replacing the language vector of \cite{Tiedemann:17}. For example, translating "This is a sentence ." to the Estonian Wikipedia domain would mean an input of "This$\vert$2wi is$\vert$2wi a$\vert$2wi sentence$\vert$2wi .$\vert$2wi"\footnote{the "$\vert$" is a special symbol in Nematus for delimiting input features.} 

Having a single language model learn several languages helps similar languages improve each others representations \cite{Tiedemann:17}. Also, they point out that this greatly alleviates the problem of sparse data for smaller languages. We expect the same effect for text domains, especially since similarity between different domains of the same languages is higher than between different languages. Moreover, similarly to the domain tag approach, the usage of many domains in one model helps bypass the over-fitting problem of smaller domains.

\subsection{Automatic domain tags}

Here we define the domain of each of the source--target sentence pair automatically. We take two different approaches to achieve the annotation.

\paragraph{Supervised approach} is done only in single domain setting. It involves assigning categories to roughly 10,000 Wikipedia articles, for which it could be done with high certainty. Assigning categories to more articles is problematic, because the categories assigned in Wikipedia can often be misleading in terms of content. Next we tag each sentence with the article category.

After tagging the sentences, we train a FastText ~\cite{fasttext1,fasttext2} classification model with default settings and apply it to classify the rest of the sentences that were not classified based on the article categories. Test/dev set sentences are tagged using the same FastText model that is used to cluster training data.

\paragraph{Unupervised approach} is applied to sentence-split data. In case of multi-domain data we still treat it as a single domain data of which we have no domain structure knowledge. In this approach, we train a model and calculate sentence vectors in an unsupervised manner using sent2vec ~\cite{sent2vec}. After that, we apply KMeans clustering to identify the clusters in the set of calculated sentence vectors. Finally, we tag each sentence with the label that it was assigned by KMeans. To find the optimal number of clusters, we create several versions with different numbers of clusters. 

To tag the test/dev set sentences, we train a FastText ~\cite{fasttext1} ~\cite{fasttext2} supervised classification model on the tagged training set.  For each of the cluster versions and for each language pair, we train a separate FastText model. The additional benefit of this kind of clustering is that each new input sentence can be efficiently assigned its cluster. Also, because of more potentially homogenous train-set clusters, the new sentence is hypothetically assigned more appropriate domain than it would be assigned in case of the pre-defined domains.

The potential benefit of the unsupervised approach over supervised approach is that it does not assume any prior knowledge of the data and thus the domain structure does not rely on potentially faulty pre-defined domain structure. This in turn allows the multi-domain translation approach to be applied to any data without the knowledge of its domain structure.


\section{Experiments with Known Domains}
\label{sctExp1}


In the experiments we use mixed-domain parallel data consisting of Europarl \cite{euparl}, OpenSubtitles2016 \cite{opsubs}, parallel data extracted from English-Estonian Wikipedia articles and some more mixed parallel corpora from the OPUS collection \cite{opsubs}. The size of the corpora is shown in Table \ref{fig:data_size}. For each corpus we use a randomly chosen and held-out test set of 3000 parallel sentences.

\begin{table} [htbp]
\centering
\begin{tabular}{| l | c | c | c |}
\hline
  \textbf{Corpus} & \textbf{Sents} & \textbf{EN tok} & \textbf{ET tok} \\ \hline \hline
  \textbf{Opensubs} & 10.32 & 83.57 & 67.56 \\ \hline
  \textbf{Europarl} & 0.644 & 17.18 & 12.82 \\ \hline
  \textbf{Wiki} & 0.135 & 2.281 & 2.089 \\ \hline
  \textbf{Other} & 7.972 & 169.9 & 143.5 \\ \hline \hline
  \textbf{Total} & 19.07 & 272.9 & 225.9\\ \hline
\end{tabular}
%
\caption{Data sizes for the training data. Number of tokens (tok) is given pre-BPE. All of the numbers are given in millions}
\label{fig:data_size}
\end{table}

\subsection{Mining Wikipedia for Translations}

Wikipedia\footnote{\url{http://www.wikipedia.org/}} itself is a big set of articles. The articles have two properties, which are extremely useful from our task point of view. Firstly, the articles have links to the articles of same topic, but in different languages, which makes it easier to find comparable data from which to extract parallel data. Secondly, each article has one or several categories attached to it. This means that hypothetically we can assign domain(s) to at least some of the articles based on these categories. 



To extract meaningful text from the Wikipedia XML dumps, we used the WikiExtractor tool\footnote{\url{https://github.com/attardi/wikiextractor}}. The data is extracted in a way that preserves article and paragraphs boundaries. The extraction is done separately for English and Estonian version.

After extracting text from the dumps, another custom-made solution is applied to detect parallel articles. The number of Wikipedia articles in English is well over 5 million whereas for Estonian it is just over 100 thousand. We keep all Estonian articles and only those English articles that have a parallel article in Estonian articles. This leaves us with roughly 70 thousand English articles. 

The parallel articles form a comparable corpus. In case of this comparable corpora we know that the articles are parallel in terms of topics but not in sentences. 
To extract parallel sentences from parallel articles, we used the LEXACC ~\cite{LEXACC} tool, which is a part of the ACCURAT 
toolkit ~\cite{ACCURAT1,ACCURAT2}.
Parallel sentence identification allows also to maintain the info of article origin, which means that direct domain assigning is possible. The identification process also assigns score to each sequence pair, which allows us to create parallel sets with different grade of purity. The optimal grade of purity produced 340 thousand parallel sentences. The size of Estonian Wikipedia in total is 2.8 million sentences. To the rest 2.5 million sentences back-translation is applied to extend the Wikipedia dataset for EN-ET direction; the back-translated sentences are also filtered based on attention weights ~\cite{attfilter} with a 50\% threshold.

\subsection{Technical Settings}

We apply BPE segmentation \cite{bpe} in a joint learning scenario, learning from the input and the output,
limiting the vocabulary to 65,000 entries. The acquired segmentation mostly corresponds to the linguistic intuition on frequent tokens (which are left intact) and medium-frequency tokens (which are split into compound parts or endings off stems); low-frequency tokens (also names, numeric tokens) are split into letters and letter pairs.

The NMT model we use is encoder-decoder with an attention mechanism \cite{bahdanau}, 
implemented in Nematus \cite{Sennrich:17}. 
All settings (like embedding size, number of recurrent layers in encoder and decoder, etc.) are kept at their default values. Batch size in experiments is 50 sequences.

\subsection{Results}

For the \textbf{Baseline} experiment we first train a baseline model on all the datasets are used, and use it for translation. Then in the \textbf{Tuned} approach for each dataset separately we fine-tune the \textbf{Baseline} model to each corpus separately.

For the comparability of the results, the number of iterations during training (800,000) and input parameters are kept equal for \textbf{Baseline}, \textbf{Tag}, \textbf{Feat}. The tuning of \textbf{Baseline} is done for additional 60,000 iterations. One iteration means one batch seen during training.


Tables~\ref{fig:ET-EN_btrans} and \ref{fig:EN-ET_btrans} show the BLEU scores \cite{bleu} and the p-values of the statistical significance of their difference for Baseline, fine-tuned baseline, domain tags, and domain feature approaches. 

\begin{table*} [htbp]
\centering
\begin{tabular}{| l | c | c | c | c |}
\hline
  \textbf{Corp} & \textbf{Baseline} & \textbf{Tuned} & \textbf{Tag} & \textbf{Feat} \\ \hline
  \textbf{Eu} & 33.0$\pm$0.3 & 35.4$\pm$0.3 & 36.2$\pm$0.3 & 37.3$\pm$0.3 \\ \hline
  \textbf{Op} & 27.9$\pm$0.6 & 28.1$\pm$0.6 & 30.5$\pm$0.6 & 30.3$\pm$0.6 \\ \hline
  \textbf{Wi} & 15.3$\pm$0.4 & 15.4$\pm$0.4 & 16.9$\pm$0.4 & 17.7$\pm$0.4 \\ \hline
\hline
  \textbf{Corp} & \textbf{Baseline} & \textbf{Tuned} & \textbf{Tag} & \textbf{Feat} \\ \hline
  \textbf{Eu} & 0.0001 \textbf{/} 0.0001 & 0.009 \textbf{/} 0.0001 & - \textbf{/} 0.0001 & 0.0001 \textbf{/} - \\ \hline
  \textbf{Op} & 0.0001 \textbf{/} 0.0001 & 0.0001 \textbf{/} 0.0001 & - \textbf{/} 0.1 & 0.1 \textbf{/} - \\ \hline
  \textbf{Wi} & 0.0001 \textbf{/} 0.0001 & 0.0001 \textbf{/} 0.0001 & - \textbf{/} 0.001 & 0.001 \textbf{/} - \\ \hline
\end{tabular}
%
\caption{BLEU scores and p-values for Estonian-English direction. \textbf{Baseline} model is trained without domain tags. \textbf{Tuned} is achieved by tuning these models with the specific corpus. \textbf{Tag} is trained with data that has domain tag prepended to each source sentence. \textbf{Feat} is trained with data that has domain embedding added as a feature to each source sequence word. p-values are given for significance against \textbf{Tag} and \textbf{Feat} respectively, separated with \textbf{/}. }
\label{fig:ET-EN_btrans}
\end{table*}

\begin{table*} [htbp]
\centering
\begin{tabular}{| l | c | c | c | c |}
\hline
  \textbf{Corp} & \textbf{Baseline} & \textbf{Tuned} & \textbf{Tag} & \textbf{Feat} \\ \hline
  \textbf{Eu} & 22.5$\pm$0.3 & 25.3$\pm$0.3 & 25.4$\pm$0.3 & 24.9$\pm$0.3 \\ \hline
  \textbf{Op} & 24.2$\pm$0.6 & 24.5$\pm$0.6 & 24.8$\pm$0.6 & 25.3$\pm$0.6 \\ \hline
  \textbf{Wi} & 11.8$\pm$0.4 & 12.1$\pm$0.4 & 12.5$\pm$0.3 & 12.8$\pm$0.4 \\ \hline
\hline
  \textbf{Corp} & \textbf{Baseline} & \textbf{Tuned} & \textbf{Tag} & \textbf{Feat} \\ \hline
  \textbf{Eu} & 0.0001 \textbf{/} 0.0001 & 0.3 \textbf{/} 0.04 & - \textbf{/} 0.04 & 0.04 \textbf{/} - \\ \hline
  \textbf{Op} & 0.01 \textbf{/} 0.001 & 0.09 \textbf{/} 0.03 & - \textbf{/} 0.06 & 0.06 \textbf{/} - \\ \hline
  \textbf{Wi} & 0.01 \textbf{/} 0.001 & 0.06 \textbf{/} 0.03 & - \textbf{/} 0.14 & 0.14 \textbf{/} - \\ \hline
\end{tabular}
%
\caption{BLEU scores and p-values for English-Estonian direction. \textbf{Baseline} model is trained without domain tags. \textbf{Tuned} is achieved by tuning these models with the specific corpus. \textbf{Tag} is trained with data that has domain tag prepended to each source sentence. \textbf{Feat} is trained with data that has domain embedding added as a feature to each source sequence word. p-values are given for significance against \textbf{Tag} and \textbf{Feat} respectively, separated with \textbf{/}. }
\label{fig:EN-ET_btrans}
\end{table*}

As we can see from the results, both of the additional domain info models perform really well. The domain tag (\textbf{Tag}) model outperforms both of its baseline (\textbf{Baseline}) and tuned (\textbf{Tuned}) counterpart in ET--EN direction. It even goes as far as exceeding the \textbf{Tuned} approach by more than 1.0 BLEU in all domains. The same holds, but even more strongly, for the version where we add the domain embedding as an input feature for each word (\textbf{Feat}).

For EN--ET direction the results do not show such strong improvements. In this direction both \textbf{Tag} and \textbf{Feat} outperform \textbf{Baseline} for all domains. However, the scoring is quite close to the \textbf{Tuned} approach with the results between \textbf{Tag} and \textbf{Feat} also being closer than in ET--EN case. All in all, the fact that the domain tagging results are essentially on-par with Tuned approach, means it is superior to the \textbf{Tuned} approach in practice because of the fact that it requires only one model rather than three.



Table~\ref{fig:ET-EN_ex1} shows an example of the \textbf{ET--EN} translations highlighting some improvements. Since the quality of \textbf{Tuned} is close to \textbf{Tag} and \textbf{Feat}, we omit it from the comparison since the differences would be highly circumstantial and would not hold much information in small scale. 

\begin{table} [htbp]
\centering
\begin{tabular}{| l | l |}
\hline
  \textbf{Src} & \textit{vastuseid saab muidugi olla ainult üks} \\
  (ET) & \textit{: lõpetada kohe igasugused} \\
  & \textit{läbirääkimised Türgiga .}\\ \hline
  \textbf{Ref} & \textit{there is , of course , only one possible} \\ 
  (EN) & \textit{response : to immediately cease all} \\ 
  & \textit{negotiations with Turkey .}\\ \hline
  \textbf{Base} & \textit{only one can only be one : stop any} \\
  (EN)& \textit{negotiations with Turkey immediately .}\\ \hline
  \textbf{Tag} &  \textit{the answer , of course , can only be} \\
  (EN) & \textit{one : stop all the negotiations with} \\ 
  & \textit{Turkey immediately .}\\ \hline
  \textbf{Feat} & \textit{there is , of course , only one answer :} \\
  (EN) & \textit{to put an end to all negotiations with} \\
  & \textit{Turkey immediately .}\\ \hline
\end{tabular}
%
\caption{An example of Europarl corpus translations from Estonian to English using the Baseline, \textbf{Tag} and \textbf{Feat} methods.}
\label{fig:ET-EN_ex1}
\end{table}

\section{Experiments with Automatic Domains}
\label{sctExp2}

Since the results on the full parallel data show that both of multi-domain approaches are on-par, or superior to the single-domain baseline, we apply the methods in a setting where we do not assume beforehand knowledge of the origin domain of source sentences. Here we take the domain tagging approach: even though domain features show better results, domain tags are more generic and compatible with any NMT architecture.

We experiment with two data settings. In the first one, we have a single heterogeneous text domain. We explore both supervised and unsupervised tagging of single text domain based on sentence vectors. 

In the second one, we have texts from several domains but we ignore the pre-specified text domains and replace them with automatic clustering based on sentence embeddings.


\subsection{Automatic single-domain tagging}








To choose the best setting for unsupervised approach, we do a small sweep for input data versions. We check for best number of clusters by training a model for each number of clusters. The input data for this is the whole Wikipedia corpus. The models are trained for 12 hours, which should be sufficient to make them diverge enough to choose the best number of clusters. We also train a regular model without data clustering for reference.


It is important to note that for this experiment a different test set was used than in the full data experiments. Thus the scores in \ref{fig:Wiki_sweep} are not comparable to scores presented earlier.

\begin{table*} [htbp]
\centering
\begin{tabular}{| l | c | c | c | c | c | c |}
\hline
  \textbf{NClust} & \textbf{C4} & \textbf{C5} & \textbf{C6} & \textbf{C8} & \textbf{C12} & \textbf{Ref}\\ \hline
  \textbf{BLEU} & 19.7 & 19.5 & 19.6 & 19.5 & 20.0 & 17.9 \\ \hline
\end{tabular}
%
\caption{BLEU scores for Unsupervised Wikipedia parameter setting.}
\label{fig:Wiki_sweep}
\end{table*}

The initial sweep indicates that the best option for the unsupervised classification is 12 clusters. Also, the 12 hours -- ~100,000 iterations are already showing the effect that domain tagging has over the regular reference approach, making other clusterings also a viable choice.

\subsubsection*{Wikipedia Translation Results}

In the final experiment, three models were trained:
\begin{itemize}
\item Supervised 5-domain source tag model
\item Unsupervised 5-domain source tag model
\item Unsupervised 12-domain source tag model
\item Regular not domain-tagged model
\end{itemize}
Unsupervised 5-domain model was included to compare the performance of supervised and unsupervised approach with the same amount of domains, giving an indication of the "goodness" of these cluster assignments. The Unsupervised 12-domain model was included to compare the performance of best unsupervised clustering and the intuitively optimal supervised clustering. Supervised 12-domain model is not presented because we were not able produce such reasonable structure from ET Wikipedia. The results are presented in \ref{fig:Wiki_final}. The models were trained for 48 hours. 

\begin{table*} [htbp]
\centering
\begin{tabular}{| l | c | c | c | c |}
\hline
  \textbf{NClust} & \textbf{Usup12} & \textbf{Usup5} & \textbf{Super} & \textbf{Ref}\\ \hline
  \textbf{BLEU} & 26.0$\pm$0.4 & 25.2$\pm$0.4 & 25.8$\pm$0.4 & 23.6$\pm$0.4 \\ \hline
  \textbf{pU12} & - & 0.01 & 0.1 & 0.0001 \\ \hline
  \textbf{pU5} & 0.01 & - & 0.03 & 0.0001 \\ \hline
  \textbf{pSup} & 0.1 & 0.03 & - & 0.0001 \\ \hline
  \textbf{pRef} & 0.0001 & 0.0001 & 0.0001 & - \\ \hline
\end{tabular}
%
\caption{BLEU scores and p-values for test on Wikipedia-only data to compare the effect of Unsupervised clustering (\textbf{Usup12, Usup5}), supervised clustering (\textbf{Super}) and no-clustering approach (\textbf{Ref}). The p-values are shown in respect to the version where the value is \textbf{-}.}
\label{fig:Wiki_final}
\end{table*}

As we see in Table~\ref{fig:Wiki_final}, the Supervised approach (\textbf{Super}) with five clusters slightly outperforms Unsupervised 5-cluster approach (\textbf{Usup5}). The best option for Unsupervised clustering (\textbf{Usup12}) performs as well as the Supervised approach. The results show that Unsupervised approach is comparable in performance to the Supervised approach, which means that at least in this setting both of the approaches are viable. Even more so, when obtaining labelled data for supervised clustering can often require a lot of additional effort, the unsupervised approach is not chained by the (lack) of pre-existing knowledge about the data.

Most important is the fact that both of the unsupervised cluster versions outperform the regular reference (\textbf{Ref}) version where sentence cluster tags were not used. This shows that the unsupervised clustering approach can potentially be used in settings that previously were viewed upon as single clusters. For example OpenSubtitles corpus could be clustered further, to improve the translations.

\subsection{Unsupervised multi-domain tagging}
\label{sctExp3}

Hinging on the fact that domain tagging approach outperformed the traditional tuning approach and on the results that unsupervised Wikipedia dataset clustering produced, the "traditional" approach of text domains should be given another look. One possible action is to cluster or sub-cluster the existing parallel data to restructure it from the domain point of view.

In addition to the results produced on wikipedia dataset, the hypothesis on why this would work, is that large text domains are probably not very homogenous. Also, different domains have probably pretty big overlap of similar sentences. This would mean that the usual approach of domain tuning or domain tagging does not achieve its true potential, because predefined domains are \textit{de facto} several domains and the same domains are actually present in other predefined domains also. 

To check for this property and its potential benefit for NMT, we cluster existing parallel sentences to $n$ clusters in the previously described unsupervised manner, train NMT models with domain tagged sentences, and finally, cluster test set sentences in a supervised manner with a supervised clustering model that is trained on the data that was obtained from unsupervised clustering. 


The training is done using Nematus with the same settings as in the initial experiment with domain tags. Firstly, we do the sweep of clusters by training 4, 8, 16, and 32 cluster versions for both EN--ET and ET--EN direction. After that we choose the version that has achieved the best BLEU scores on the dev sets for both of the directions and train it for the same time as in the initial domain tag experiment with full data. 

\subsubsection*{Results of unsupervised multi-domain tagging}

To evaluate the model performance, we train supervised FastText classification models on the tagged training data. We apply these models on the test/dev sets to classify the sentences. This means that each of the sets -- Opensubs, Europarl, and Wiki -- gets actually tags from several clusters, depending on which cluster the FastText model assigns to each of the sentences. This means that for each source test set we create four different versions, each for cluster numbers 4, 8, 16, and 32. 

The initial parameter sweep shows that the best option is 16 clusters for both EN--ET \ref{fig:Full_sweep_ENET} and ET--EN \ref{fig:Full_sweep_ETEN} directions across all test sets. Hence the final models were both trained with 16 clusters.

\begin{table} [htbp]
\centering
\begin{tabular}{| l | c | c | c | c |}
\hline
  \textbf{Corp} & \textbf{C4} & \textbf{C8} & \textbf{C16} & \textbf{C32} \\ \hline
  \textbf{Eu} & 4.13 & 3.19 & \textbf{5.94} & 4.17 \\ \hline
  \textbf{Op} & 9.41 & 9.36 & \textbf{10.80} & 10.62 \\ \hline
  \textbf{Wi} & 1.09 & 0.94 & \textbf{1.31} & 0.81 \\ \hline
\end{tabular}
%
\caption{BLEU scores for English-Estonian direction sweep. The model is trained on parallel data that is tagged in unsupervised manner using sent2vec + Kmeans clustering. The dev sets are clustered based on this tagged data using FastText. The best scores for each corpus are presented in \textbf{bold}.}
\label{fig:Full_sweep_ENET}
\end{table}

\begin{table} [htbp]
\centering
\begin{tabular}{| l | c | c | c | c |}
\hline
  \textbf{Corp} & \textbf{C4} & \textbf{C8} & \textbf{C16} & \textbf{C32}\\ \hline
  \textbf{Eu} & 20.48 & 19.88 & \textbf{20.82} & 18.43 \\ \hline
  \textbf{Op} & 20.05 & 19.54 & \textbf{20.17} & 20.01 \\ \hline
  \textbf{Wi} & 4.61 & 4.38 & \textbf{5.50} & 4.32 \\ \hline
\end{tabular}
%
\caption{Test set BLEU scores for Estonian-English direction sweep. The model is trained on parallel data that is tagged in unsupervised manner using sent2vec + Kmeans clustering. The dev sets are clustered based on this tagged data using FastText. The best scores for each corpus are presented in \textbf{bold}.}
\label{fig:Full_sweep_ETEN}
\end{table}

In table \ref{fig:ClustrStructEN} is shown the OpenSubs test sets cluster structure. The test sets are tagged using FastText models trained on tagged train set. We can see that different train set clusters produce different granularity in test sets also. For \textbf{C4, C8} the OpenSubs structure is similar, same holds for other test sets. \textbf{C16} vs \textbf{C8} however shows a significant difference in test set clustering. Here we see that OpenSubs, which based on content is probably not homogeneous domain, is separated quite granularly in \textbf{C16}, producing 3--4 main sub-domains. In \textbf{C32} the test set is clustered even further, but based on sweep scores, it could be said that the achieved clustering is already too granular.

\begin{table} [htbp]
\centering
\begin{tabular}{| l | c | c | c | c | c | c |}
\hline
  \textbf{Corp} & \textbf{N1} & \textbf{N2} & \textbf{N3} & \textbf{N4} & \textbf{N5} & \textbf{N6}\\ \hline
  \textbf{C4} & 2921 & 29 & - & - & - & - \\ \hline
  \textbf{C8} & 2907 & 43 & - & - & - & - \\ \hline
  \textbf{C16} & 1331 & 1015 & 398 & 181 & 18 & 7 \\ \hline
  \textbf{C32} & 1137 & 828 & 356 & 293 & 241 & 71 \\ \hline
\end{tabular}
%
\caption{Cluster structure of FastText tagged English OpenSubs test sets. The test sets are clustered based on tagged train data. The clusters are numbered left to right based on size. Here only top 6 clusters are shown. For C32 $N7=11$, $N8=6$, $N9=3$, $N10=2$, $N11=2$. Test set structures for Estonian sets are similar.}
\label{fig:ClustrStructEN}
\end{table}  

Considering that our OpenSubs cluster is 10 million sentence pairs in size, we can say that \textbf{C16} finds 5 significant sub-domains and one less significant sub-domain inside it. This shows that, at least from sentence vectorizing point of view, there exists more than one domain inside OpenSubs, and similarly in other domains.

\begin{table*} [htbp]
\centering
\begin{tabular}{| l | c | c | c | c | c | c | c | c |}
\hline
  \textbf{Corp} & \textbf{N1} & \textbf{N2} & \textbf{N3} & \textbf{N4} & \textbf{N5} & \textbf{N6} & \textbf{N7} & \textbf{N8} \\ \hline
  \textbf{Train} & 4859672 & 4444177 & 3704753 & 2767889 & 1407228 & 822225 & 711526 & 134301 \\ \hline
  \textbf{Corp} & \textbf{N9} & \textbf{N10} & \textbf{N11} & \textbf{N12} & \textbf{N13} & \textbf{N14} & \textbf{N15} &  \textbf{N16}\\ \hline
\textbf{Train} & 114778 & 40260 & 22004 & 18165 & 10298 & 9585 & 2492 & 646 \\ \hline
\end{tabular}
%
\caption{Cluster structure of KMeans tagged English train set for \textbf{C16}. The clusters are numbered left to right based on size. Train set structure for Estonian is similar.}
\label{fig:ClustrTrainEN}
\end{table*}  

When looking at the number of clusters present in Table~\ref{fig:ClustrStructEN}, one could notice that the clusters present is less than number of clusters defined. It should be kept in mind that we have 3 main text sources in training set and fourth mixed-corpus which could be divided into 5-6 parts, so 8-9 text domains in total. Also, some sentences are quite distinct from the others based on full train set cluster structure as we can see from the train set structure of \textbf{C16} in Table~\ref{fig:ClustrTrainEN}. The clustering and its structure is probably interesting aspect to look into in future work.

The final results, where the 16 cluster models were trained for the same amount of iterations as in the initial full data experiments, are presented in \ref{fig:EN-ET_full} and \ref{fig:ET-EN_full} for EN--ET and ET--EN language pairs respectively.

\begin{table*} [htbp]
\centering
\begin{tabular}{| l | c | c | c | c |}
\hline
  \textbf{Corp} & \textbf{Baseline} & \textbf{Tuned} & \textbf{Tag} & \textbf{Unsup} \\ \hline
  \textbf{Eu} & 22.5$\pm$0.3 & 25.3$\pm$0.3 & 25.4$\pm$0.3 & 24.5$\pm$0.3 \\ \hline
  \textbf{Op} & 24.2$\pm$0.6 & 24.5$\pm$0.6 & 24.8$\pm$0.6 & 24.6$\pm$0.6 \\ \hline
  \textbf{Wi} & 11.8$\pm$0.4 & 12.1$\pm$0.4 & 12.5$\pm$0.3 & 11.1$\pm$0.4 \\ \hline
\hline
  \textbf{Corp} & \textbf{Baseline} & \textbf{Tuned} & \textbf{Tag} & \textbf{Unsup} \\ \hline
  \textbf{Eu} & 0.0001 \textbf{/} 0.0001 & 0.3 \textbf{/} 0.03 & - \textbf{/} 0.004 & 0.004 \textbf{/} - \\ \hline
  \textbf{Op} & 0.01 \textbf{/} 0.03 & 0.09 \textbf{/} 0.4 & - \textbf{/} 0.2 & 0.2 \textbf{/} - \\ \hline
  \textbf{Wi} & 0.01 \textbf{/} 0.01 & 0.06 \textbf{/} 0.005 & - \textbf{/} 0.0001 & 0.0001 \textbf{/} - \\ \hline
\end{tabular}
%
\caption{Test set BLEU scores and p-values for English-Estonian direction. \textbf{Baseline} model is trained without domain tags. \textbf{Tuned} is achieved by tuning these models with the specific corpus. \textbf{Tag} is trained with data that has domain tag prepended to each source sentence. \textbf{Unsup} is trained with data that has domain tags assigned to each sentence in an previously described unsupervised manner. p-values are given for significance against \textbf{Tag} and \textbf{Unsup} respectively, separated with \textbf{/}. }
\label{fig:EN-ET_full}
\end{table*}

\begin{table*} [htbp]
\centering
\begin{tabular}{| l | c | c | c | c |}
\hline
  \textbf{Corp} & \textbf{Baseline} & \textbf{Tuned} & \textbf{Tag} & \textbf{Unsup} \\ \hline
  \textbf{Eu} & 33.0$\pm$0.3 & 35.4$\pm$0.3 & 36.2$\pm$0.3 & 36.0$\pm$0.3 \\ \hline
  \textbf{Op} & 27.9$\pm$0.6 & 28.1$\pm$0.6 & 30.5$\pm$0.6 & 30.2$\pm$0.6 \\ \hline
  \textbf{Wi} & 15.3$\pm$0.4 & 15.4$\pm$0.4 & 16.9$\pm$0.4 & 16.0$\pm$0.4 \\ \hline

\hline
  \textbf{Corp} & \textbf{Baseline} & \textbf{Tuned} & \textbf{Tag} & \textbf{Unsup} \\ \hline
  \textbf{Eu} & 0.0001 \textbf{/} 0.0001 & 0.009 \textbf{/} 0.01 & - \textbf{/} 0.3 & 0.3 \textbf{/} - \\ \hline
  \textbf{Op} & 0.0001 \textbf{/} 0.0001 & 0.0001 \textbf{/} 0.0001 & - \textbf{/} 0.1 & 0.1 \textbf{/} - \\ \hline
  \textbf{Wi} & 0.0001 \textbf{/} 0.004 & 0.0001 \textbf{/} 0.009 & - \textbf{/} 0.01 & 0.01 \textbf{/} - \\ \hline
\end{tabular}
%
\caption{Test set BLEU scores and p-values for Estonian-English direction. \textbf{Baseline} model is trained without domain tags. \textbf{Tuned} is achieved by tuning these models with the specific corpus. \textbf{Tag} is trained with data that has domain tag prepended to each source sentence. \textbf{Unsup} is trained with data that has domain tags assigned to each sentence in an previously described unsupervised manner. p-values are given for significance against \textbf{Tag} and \textbf{Unsup} respectively, separated with \textbf{/}.}
\label{fig:ET-EN_full}
\end{table*}

The results show that the unsupervised clustering approach performs similarly with the pre-defined tag version. The results are evidence that the unsupervised tagging approach can serve as a viable alternative to the traditional pre-defined domain approach. Our hypothesis is that this is caused by the pre-defined domains being less homogenous in content than the unsupervised clustered "domains". However, this hypothesis should be investigated further to assert its existence and magnitude. Also, since the clustering approach is pretty much applied out-of-the-box, then improved clustering could provide considerable improvements. 

All-in-all, taking into consideration the fact that unsupervised approach allows new sentences to be translated with potentially more appropriate domain assigned to them, the unsupervised tagging approach can be seriously considered as the go-to approach for multi-domain translation models.

\section{Discussion}
\label{sctAnalysis}

The results from the experiments - EN--ET and ET--EN direction parallel translation, Wikipedia data translation, and unsupervised sentence tagging - show that both of the two chosen multi-domain approaches outperform regular approach of uniform translation and domain-tuning. 

This indicates the hypothesis that the parameter sharing effect discussed in Google's zero-shot article would benefit domain translation holds. The translation scores even outperform domain-tuning approach, which could be explained by the same parameter sharing. In tuning we tune the model to translate sentences characteristic to the model we are tuning to. This means that domain-characteristic sentences get translated really well. On the other hand, the not-so-characteristic sentences get neglected. The parameter sharing effect of the multi-domain approach helps negate the negative effect by the support of other domains while still learning to more effectively represent each domain by the additional domain info.

Furthermore, the results indicate that adding domains as an input feature can have even stronger effect on the translation scores. This shows that concatenating the domain feature embedding with word embedding at each timestep - basically remembering the source domain equally throughout the sequence improves model performance. This could be explained by the fact that in tag prepending case, the neural net may "forget" for longer sequences what the input tag was, making the effect of it weaker. 

The results also show that for highly quality dependent settings the domain feature concatenation with word embedding is the more suitable option. However, the differences in scores are not drastically different from the domain tag prepending. This means that for the sake of data simplicity, model simplicity and efficiency the tag prepending approach could prove more reasonable of the two for in-production settings.

Finally, the performance of unsupervised domain tagged model indicates that there is grounds to substitute the pre-defined domain approach with automatically assigned domain approach. The unsupervised certainly serves as an improvement in less homogenous single domain settings, where the effect of the detection of underlying "domains" was shown on the example of Wikipedia.

No less important are the facts that the unsupervised tagging approach ensures better domain assignment to each new sentence and can efficiently incorporate new data from various small domains to fortify each of the learned "domain" (clusters). 

It has to be taken into account that the unsupervised clustering performed in these experiments is applied basically in out-of-the-box manner, which means that domain assignments can be improved and thus the translation scores should also improve.

\section{Conclusions}
\label{sctConclusions}

In this article we tested two approaches to improve multi-domain neural translation. One approach involves prepending domain tags to source sentences, the other adding domain embeddings as an input feature to each source sentence word. We showed that both ways of adding domain information to source sentences in bilingual neural translation improves translation scores considerably compared to both regular baseline translation and fine-tuning. These improvements in source sentence tagging case can be obtained with mere data manipulation.

We also showed that the domain tagging approach can be successfully coupled with unsupervised sentence clustering to add a "domain dimension" to a previously single-domain corpus. This approach produces better results as opposed to using the corpus as a single domain. The results indicate that unsupervised or semi-supervised training data clustering can be effectively used to improve neural machine translation.

Finally, to bring the two experiments together, we apply unsupervised domain tagging to full parallel data and show that it can serve as a viable alternative to the pre-defined domain approach.

For future work the clustering in fully unsupervised tagging approach should be improved to see if this gives a visible improvement in translation scores.

Secondly, a more comprehensive sweep on number of clusters should be done. It would be interesting to see for how many clusters the effect still persists. This however would need more extensive computational resources and should probably be done with some model dataset.

The differences of the two approaches - source sentence tagging and adding domain info as an input feature - deserve to be looked into more deeply. More precisely, the result profiles of the two in different cases of domain granularity.

Finally, in this work domains are still treated as nominal values; it would be interesting to explore the estimation of domain embeddings at translation time as continuous values.

\section*{Acknowledgements}

This work was supported by the Estonian Research Council grant no. 1226. Also we are grateful for the anonymous reviews and their help in improving this article.

\bibliographystyle{eamt18}
\bibliography{refs}

\end{document}